\documentclass[11pt,a4paper]{article}
\usepackage[hyperref]{naaclhlt2019}
\usepackage{times}
\usepackage{latexsym}
\usepackage{amsmath}
\usepackage{cleveref}
\usepackage{siunitx}
\usepackage{booktabs}
\usepackage{multirow}
\usepackage{chngpage}
\usepackage{graphicx}
\usepackage{xspace}
\usepackage{url}

\aclfinalcopy

\newcommand{\dataset}{{{{WikiDocEdits}}}\xspace}

\newcommand\eth{$^{\diamondsuit}$}
\newcommand\ms{$^\spadesuit$}
\newcommand\ethmsr{$^{\diamondsuit,*}$}
\newcommand\aspace{\hspace{.75em}}

\title{Text Editing by Command}

\author{
Felix Faltings\ethmsr\aspace
Michel Galley\ms\aspace
Gerold Hintz\ms\aspace
Chris Brockett\ms\aspace \\
{\bf Chris Quirk}\ms\aspace
{\bf Jianfeng Gao}\ms\aspace
{\bf Bill Dolan}\ms\aspace \\
\eth Department of Computer Science, ETH Zürich \aspace\ms Microsoft Research\\
{\tt fafelix@student.ethz.ch}\\
{\tt \{mgalley,gehint,chrisbkt,chrisq,jfgao,billdol\}@microsoft.com}
}

\date{}

\begin{document}
\maketitle
 \begin{NoHyper}
{\let\thefootnote\relax\footnotetext{* Work done at Microsoft Research.}}
\end{NoHyper}

\begin{abstract}
 A prevailing paradigm in neural text generation is one-shot generation, where text is produced in a single step. 
 The one-shot setting is inadequate, however, when the constraints the user wishes to impose on the generated text are dynamic, especially when authoring longer documents. 
 We address this limitation with an interactive text generation setting in which the user interacts with the system by issuing commands to edit existing text. 
 To this end, we propose a novel text editing task, and introduce \dataset, a dataset of single-sentence edits crawled from Wikipedia. 
 We show that our Interactive Editor,  a transformer-based model trained on this dataset, outperforms baselines and obtains positive results in both automatic and human evaluations. We present empirical and qualitative analyses of this model's performance.\footnote{All our code (including code to recreate our data) and pre-trained models will be made available at: \url{ http://microsoft.com/research/project/interactive-document-generation}}
\end{abstract}

\section{Introduction}\label{sec:intro}

A long-standing goal of natural language processing (NLP) research has been to generate long-form text \citep{Lebowitz1985StorytellingAP, Fan2018HierarchicalNS, rashkin2020plotmachines}. 
Recent large generative language models such as GPT-2 \citep{Radford2019LanguageMA}, and  GPT-3 \citep{Brown2020LanguageMA}, demonstrate an impressive ability to generate fluent text, but their outputs are difficult to control beyond a prompt, and they manifest a tendency to hallucinate facts \citep{Wiseman2017ChallengesID}. Much recent work has thus focused on making such models more controllable \citep{Keskar2019CTRLAC,Hu2017TowardCG,Zhang2020POINTERCT,Dathathri2020PlugAP}, and factually grounded \citep{Guu2020REALMRL,Liu2018TabletotextGB}. 

\begin{figure}[t]
    \centering
    \includegraphics[width=0.5\textwidth]{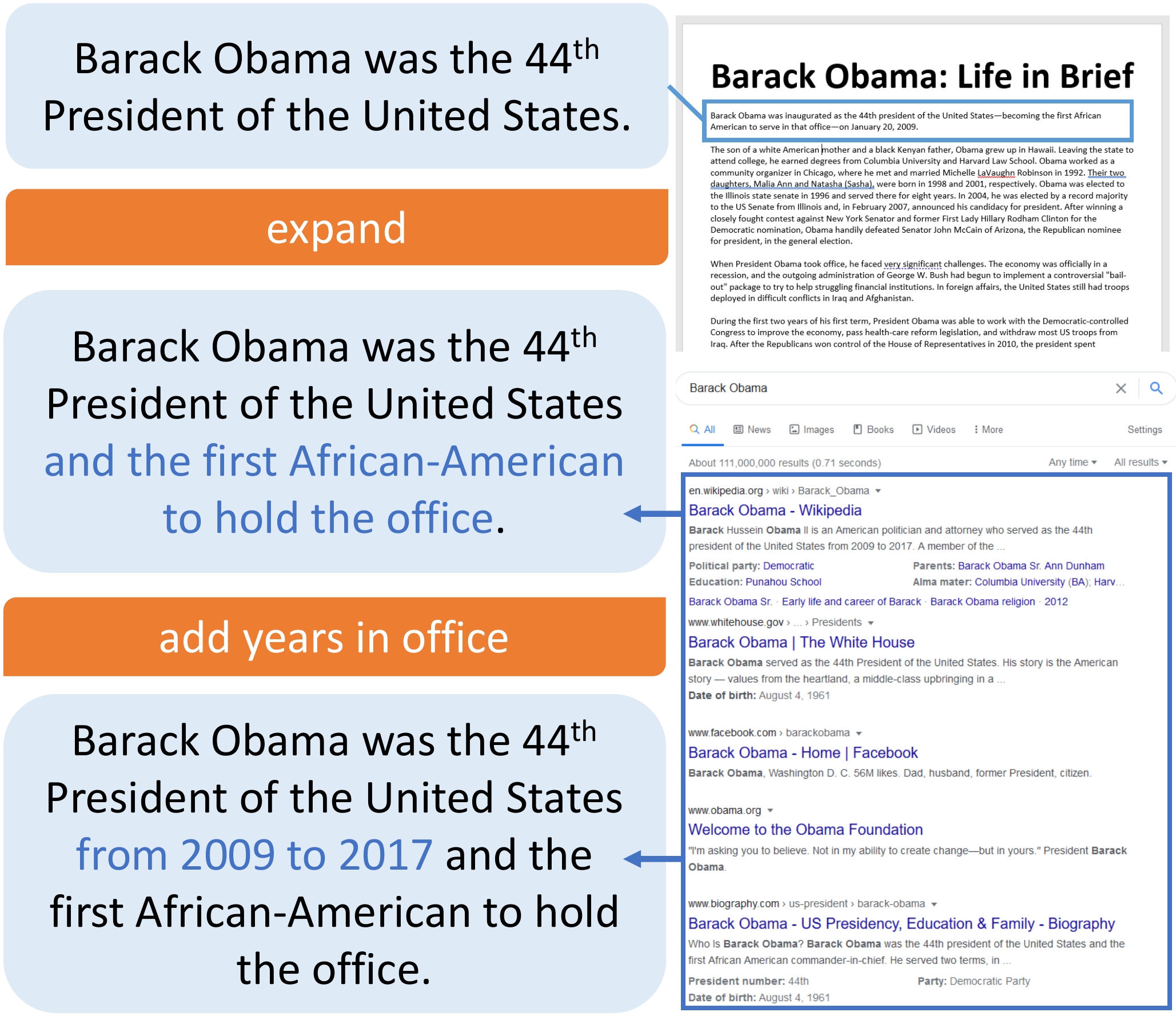}
    \caption{An illustration of our interactive text generation setting. This is an example generated by our model. The blue panels represent the text being edited, taken from the document shown on the right. The orange panels represent user edit commands. The model grounds edits in query results from a commercial search engine.}
    \label{fig:task_illustration}
\end{figure}

Most such work only considers a \textit{one-shot} generation setting. Given a set of inputs, which may be a prompt, a control code \citep{Keskar2019CTRLAC}, or a table of data \citep{Liu2018TabletotextGB} for example, the system generates text in a single step. 
Humans, though, often produce text through an evolutionary process involving multiple draft-edit cycles. This is not simply because they make mistakes when writing, but because they may require multiple iterations to help them shape and even make sense of what they want to express \citep{PirolliTheSP}. For example, consider a user writing an article about Barack Obama. They might start with a simple sentence such as ``Barack Obama was the 44th President of the United States''. Next, they may wish to expand on that sentence, adding information, or rephrasing it to integrate it better with the text. 
Replicating this process in software will mean allowing users to adjust their requirements in response to model outputs. Even an error-free system that meets all of a user's initial requirements does not obviate the need for iteration, since those constraints are themselves dynamic.

The purpose of this paper is to bring into view the task of controllable text editing, as a step beyond \textit{one-shot} generation towards \textit{interactive} document generation.  
A full interactive document generation system will likely comprise multiple components, possibly including one-shot generation to create a first draft. Editing is crucial to interactivity because it allows users to change previously generated text to fit their dynamic constraints. 
This is a stateful operation, where the state is the current version of the document, as opposed to stateless recasting of text from scratch using a one-shot model. 
While services like Grammarly or MS Word already offer rewriting suggestions, they mainly focus on small edits, such as paraphrases \citep{Gupta2018ADG}. In this work, we are interested in a broader range of edits, particularly those that add or remove content, or change the meaning of text. \Cref{fig:task_illustration} illustrates this editing setting with an example from our trained model, where a user produces a sentence about Barack Obama over multiple edits.

In sum, we make the following contributions:
We introduce a challenging new text editing task, wherein a model must learn to edit text in response to a user command, while drawing on grounding to avoid problems of hallucination \citep{Wiseman2017ChallengesID}. To accompany this task, we release an open-source dataset of sentence-level edits extracted from Wikipedia, including editor comments, which we leverage as natural language commands, together with pre-retrieved grounding documents. We show that a transformer-based editing model trained on our data outperforms ``parrot" and GPT-2 baselines, and obtains competitive results compared to gold-standard edits in human evaluations. We then perform an empirical analysis of our model's performance, showing the importance of the command and grounding, and the varying difficulty of edits in our dataset. 

\section{Text Editing Task}\label{sec:methods}
We now formalize our text editing task. Let $D$ be a document, $q$ a user command\footnote{This notation reflects that the edit command is analogous to a query in a retrieval or QA setting in that it expresses a form of user intent.}, and $\mathcal{G}$ some appropriate form of grounding. Moreover, let $D'$ be an edited version of $D$. Then our task is, given a dataset of edits $\mathcal{D}=\{(D_0, q_0, \mathcal{G}_0, D'_0), ...,(D_N, q_N, \mathcal{G}_N, D'_N)\}$, learn to produce document $D'$, given $D$, $q$, and $\mathcal{G}$.

Note that while previous work on text editing usually only considers $D$ as input, we include both a form of control $q$ and grounding $\mathcal{G}$. The command is needed because otherwise the type of edit to be made is undefined, while the grounding provides external knowledge needed to make an edit.

In our specific instance of this task, we will only consider sentence-level edits. More formally, we consider edits $D \xrightarrow{} D'$, where $D$ and $D'$ differ only on a single sentence $s\in D$, respectively $s'\in D'$. While, in general, edits can vary in complexity from document-level to character-level changes, sentences are a natural way to break down text into relatively independent units of meaning, so it makes sense to edit text one sentence at a time. More complex, document-level edits can be seen as a composition of multiple sentence-level edits.

Additionally, we will consider user commands $q$ written in natural language, e.g., ``add years in office''. The command could also take other forms, such as a categorical variable, but natural language allows for the greatest flexibility in specifying what the edit should accomplish. Moreover, natural language commands are a good fit for our model, which we will initialize with pretrained language model weights. For similar reasons, we will also consider corpora of text snippets as our grounding $\mathcal{G}$. Alternatively, the grounding could also consist of structured data such as tables or graphs. In a real user scenario this grounding might be supplied by the user, or retrieved on the fly. For our dataset, we pre-retrieve groundings by querying a commercial search engine.

\section{Data}\label{sec:data}
To accompany our text editing task we present a novel dataset of nearly 12 million sentence-level edits, \dataset. These edits were extracted from the revision histories in the February 1\textsuperscript{st} 2020 dump of English Wikipedia.\footnote{Downloadable from \href{https://dumps.wikimedia.org/}{https://dumps.wikimedia.org/}}

For a given Wikipedia page, a revision consists of a source and target text, corresponding to the old and new versions of the page. Each revision is also accompanied by an editor comment, which we will use as a proxy for the user command. For a given revision, we split the source and target texts into sentences and then attempt to match the sentences between source and target. For efficiency, we only look at a $k$-sentence neighborhood. Unmatched sentences are candidates for edits. A source sentence $s$ and target sentence $t$ form an edit pair $s\xrightarrow{}t$ if $f(s,t)>\epsilon$, where $f$ is sentence-level BLEU\footnote{We use BLEU-4 in all experiments of this paper.} without smoothing and $\epsilon = 0.1$ in our case. If an unmatched source sentence does not form an edit pair with any target sentence, we consider it to be a sentence deletion. This can also be thought of as matching to an empty sentence. We identify sentence insertions in an analogous manner. Importantly, we only consider revisions that contain a single sentence-level edit. Otherwise, the editor comment that accompanies each revision may only describe one of the possibly many sentence-level edits. See \cref{sec:appendix_data_pipeline} for a more detailed description of our processing pipeline.

\subsection{Grounding}
We retrieve grounding snippets for the edits in our dataset by querying a commercial search engine.
In order to formulate a query for a given edit, we combine the relevant page and section titles with keywords\footnote{see \cref{sec:appendix_grounding_search_keywords} for details on how we identify keywords} from the target sentence. While the target sentence is not available at test time, we make the assumption that in a real user scenario the relevant grounding would be provided by the user. 

We retrieve the top 200 returned web page results and only keep the preview snippets returned by the search engine as the grounding corpus.\footnote{We also experimented with retrieving and parsing the HTML pages from the search but this did not lead to better end-to-end performance than just using the snippets.}

Because Wikipedia, as well as several clones, often appear in search engine results, we check for $4$-gram overlap between the target sentence and each grounding snippet, removing any snippet with more than $50\%$ overlap.
Finally, we rerank\footnote{see \cref{sec:appendix_grounding_reranking} for details on reranking} the retrieved snippets using an information extraction score, and merge the ranked snippets to take the first $N=512$ tokens. 

\subsection{Data Analysis}\label{sec:data_analysis}
We now provide an overview of our dataset. From $667$ dump files in the February 1\textsuperscript{st} dump of Wikipedia, we extract 11,850,786 edits, and take a $1\%$ sample of 118,818 edits to run our analyses. \Cref{tab:data_summary_stats} presents summary statistics for our data, and in the following, we break down the edits by edit type, analyze the quality of the retrieved grounding, and present some examples.

\begin{table}%
    \centering
    \begin{tabular}{@{}l*{4}{r}@{}} \toprule
        Statistic & \multicolumn{3}{c}{Percentiles} & Mean \\ \cmidrule(r){2-4}
         & \multicolumn{1}{c}{25\%} & \multicolumn{1}{c}{50\%} & \multicolumn{1}{c}{75\%} & \\ \midrule
        Sentence length & 16 & 23 & 31 & \num{25.25187} \\
        Diff length & 2 & 3 & 9 & \num{7.26948} \\
        Comment length & 2 & 3 & 7 & \num{5.19997} \\
        \bottomrule
    \end{tabular}
    \caption{Summary statistics of \dataset. All statistics were computed on a $1\%$ subsample of the data. Lengths reported in number of words. The diff length corresponds to the number of words, inserted or deleted, affected by a given edit.}
    \label{tab:data_summary_stats}
\end{table}

\paragraph{Fluency and Content Edits}
We are interested in the distribution of different edit types within our dataset. In particular, we want to distinguish between fluency edits, which only affect the grammar or structure of a sentence, and content edits, which change the meaning of a sentence. We can lean on previous work to categorize edits on Wikipedia. \citet{Yang2017IdentifyingSE} create 13 edit intention categories, and train a classifier to label revisions according to the categories. We apply their classifier to our data, and group their 13 categories into ``fluency'', ``content'', or ``other'' edits, as reported in \cref{tab:data_intention_breakdown_grouped}. With the caveat that the edits were labelled automatically using a trained classifier, we see that, while fluency edits make up the majority of the edits in our data, a large proportion are content edits.

\begin{table}%
    \centering
    \begin{tabular}{@{}p{0.2\linewidth} p{0.5\linewidth}r@{}} \toprule
         Group & Labels & \%Edits \\ \midrule
         Fluency & Refactoring, Copy-editing, Wikification, Point-of-view & \num{56.99769311951882} \\
         Content & Fact-update, Simplification, Elaboration, Verifiability, Clarification & \num{24.76580960297373} \\
         Other & Unlabeled, Vandalism, Process, Disambiguation, Counter-vandalism & \num{26.649606335982664} \\
         \bottomrule
    \end{tabular}
    \caption{Breakdown of edits by grouped intention labels. See Table \ref{tab:data_intention_breakdown} in the appendix for a breakdown by intention label instead of group. The percentages do not total $100$ because edits can have multiple labels.}
    \label{tab:data_intention_breakdown_grouped}
\end{table}

\paragraph{Coverage Analysis}

We are also interested in knowing how well edits in the data are covered by the inputs (i.e. $D,s,q$, or $\mathcal{G}$), where an edit is well covered if the information necessary to produce the edit appears somewhere in the inputs. To measure coverage we use word recall: how many words that were inserted in an edit also appear in the grounding? However, because simple recall fails to account for synonyms, or the context in which words appear, we use the BERTScore \citep{Zhang2020BERTScoreET} recall. This allows for fuzzy matching between BERT embeddings instead of requiring exact word matches. We also use idf scores to weigh words, since we are mostly interested in covering rare words, which are more likely to be meaning-carrying. We can define the BERT recall, $R_{\text{BERT}}$, for a sentence edit $s\xrightarrow{}s'$, with respect to some text corpus $\mathcal{C}$ as
$$
\frac{\sum_{w\in s'\backslash s}\text{idf}(w)\cdot\text{max}_{w'\in\mathcal{C}}\text{BERT}(w)^T\text{BERT}(w')}{\sum_{w\in s'\backslash s}\text{idf}(w)},
$$
where $s'\backslash s=\{w\in s'| w\notin s\}$, and idf($w$) are the inverse document frequency scores computed on a random sample of $500K$ Wikipedia pages.

\Cref{tab:coverage_stats} reports the coverage statistics for our subsample of the data. We used an uncased BERT base model to compute the embeddings. The first row reports the coverage of the target by all of the inputs, namely the command, grounding, context, and source sentence. The second row shows the coverage by the grounding alone. Note that, even with just the grounding,  coverage is already fairly high. Finally, the last row presents the coverage by the command alone, which shows that it also provides grounding.

\begin{table}%
    \centering
    \begin{tabular}{@{}l*{4}{r}@{}} \toprule
        Coverage corpus & \multicolumn{3}{c}{Percentiles} & Mean \\ \cmidrule(r){2-4}
         & \multicolumn{1}{c}{25\%} & \multicolumn{1}{c}{50\%} & \multicolumn{1}{c}{75\%} & \\ \midrule
        All Inputs & \num{0.6343951169453976} & \num {0.7387174362458746} & \num{0.8262425039502773} & \num{0.631577145268652} \\
        Grounding & \num{0.4785501482317265} & \num{0.6004798710346222} & \num{0.719543606042862} & \num{0.508560105017864} \\
        Comment & \num{0.20428421610392808} & \num{0.2946273874871878} & \num{0.43981297823282184} & \num{0.2733873583738252} \\
        \bottomrule
    \end{tabular}
    \caption{$R_{\text{BERT}}$ statistics of inserted words for edits in \dataset. All statistics were computed on a $1\%$ subsample of the data. The BERT embeddings used to compute $R_{\text{BERT}}$ were produced using a pretrained BERT base model. The idf weights were computed from a sample of 500,000 Wikipedia pages. Each row represents a different recall when considering a different coverage corpus $\mathcal{C}$.}
    \label{tab:coverage_stats}
\end{table}

\paragraph{Examples}

\begin{table*}[hbt]
\begin{adjustwidth}{-.0in}{-.0in}
    \centering
    \scalebox{0.965}{
    \begin{tabular}{@{}*{2}{p{0.5\linewidth}}@{}} \toprule
        Source sentence &  Target sentence \\ \midrule
        For decades to follow, the movie was aired in the United States on or near Easter. &  For decades to follow, the movie was aired in the United States on or near Easter \textbf{, although today with the Turner cable networks now holding the television rights, the film is generally shown during the summer and Christmas seasons}. \\ \midrule
        After the execution-style killings that \textbf{inadvertenty} led to the deaths of Frank Castle's family, Russo was hired by Bruno Costa to assassinate another assassin who had failed to kill Frank Castle as well. & After the execution-style killings that \textbf{inadvertently} led to the deaths of Frank Castle's family, Russo was hired by Bruno Costa to assassinate another assassin who had failed to kill Frank Castle as well. \\ \midrule
        He married Margaret Frances Prowse Shaw in Sydney in \textbf{1874} . & He married Margaret Frances Prowse Shaw in Sydney in \textbf{1871} . \\ \midrule
        Entitled "It Feels Like Home (Re Invented) Tour 2011", it \textbf{contained his songs and} remakes of Alliage hits. & Entitled "It Feels Like Home (Re Invented) Tour 2011", it \textbf{included many} remakes of Alliage hits \textbf{as well as some of his newer songs}. \\
        \bottomrule
    \end{tabular}
    }
    \caption{Example edits from \dataset. The edited portions are highlighted in bold.}
    \label{tab:edit_examples}
\end{adjustwidth}
\end{table*}

\Cref{tab:edit_examples} presents some examples from our data. These were chosen to illustrate a variety of phenomena. The first example shows an elaboration edit, appending new information to the end of a sentence. The second example is a simple typo fix, while the third is changing a fact. Finally, the last example is a more complex edit involving insertion and deletion to reword and clarify a sentence. We can see that there is a large variety of edits in our dataset. See \cref{tab:appendix_edit_examples} in the appendix for more examples, including comments.

\section{Model}

\begin{figure}[t]
    \centering
    \includegraphics[width=0.4\textwidth]{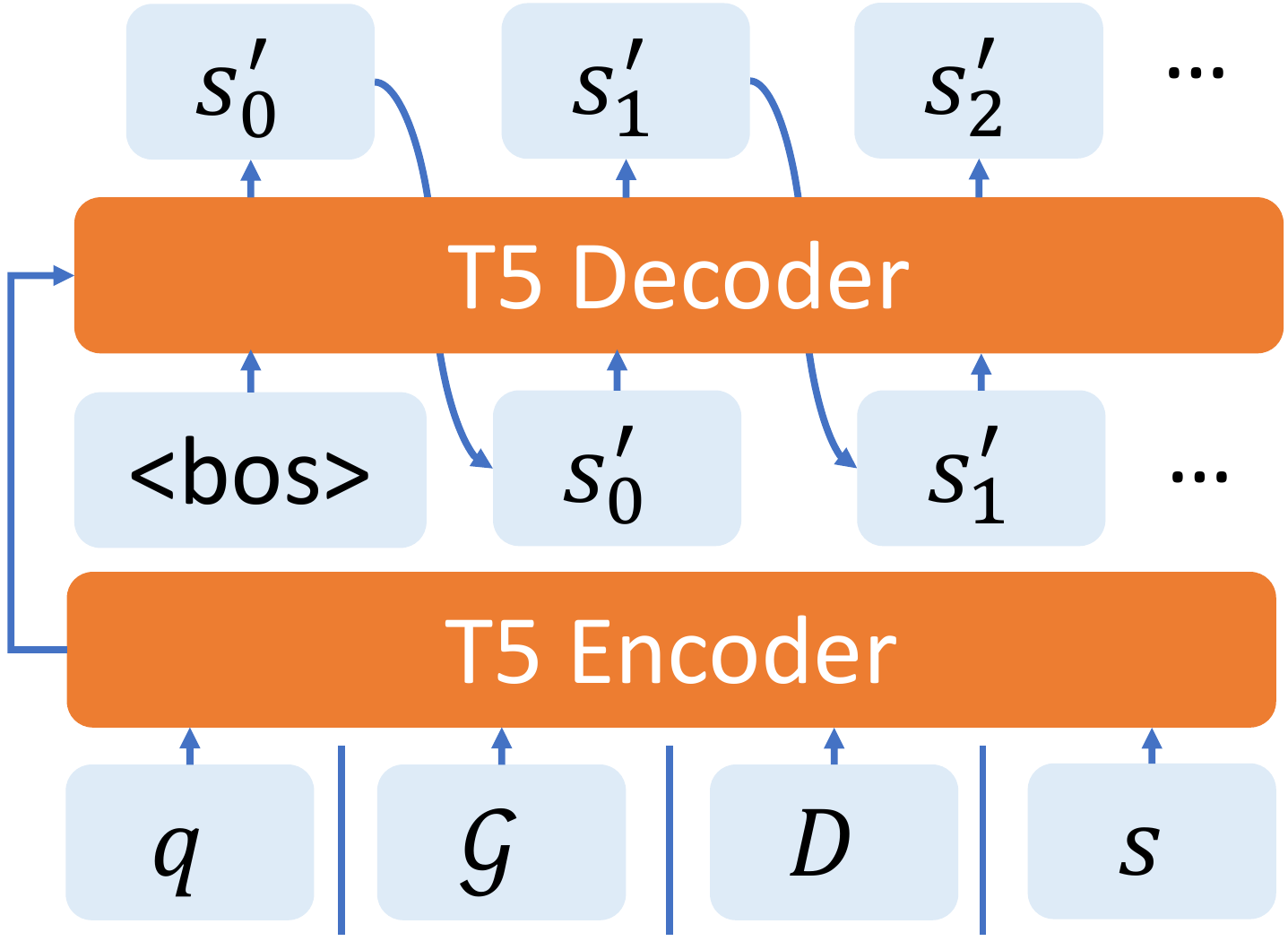}
    \caption{An illustration of our model. The inputs to the encoder are sequences of tokens separated by $\langle$ sep$\rangle$ tokens, represented by the vertical bars in the figure.}
    \label{fig:model}
\end{figure}

We formalize our model, which we refer to as Interactive Editor, as a standard auto-regressive sequence to sequence model. Because our data only contains single-sentence edits, we assume that the sentence to be edited in the source document is given as an input to the model.

Given a source sentence $s\in D$, the context around $s$, which we will refer to as $D$ by abuse of notation, a user command $q$, a grounding corpus $\mathcal{G}$, and a candidate target sentence $s'$, the model, $f$, computes
\begin{align*}
f(s,s',D,q,\mathcal{G}) &= P(s'|s,D,q,\mathcal{G}) \\
&= \prod_iP(s'_i|s'_{<i},s,D,q,\mathcal{G}),
\end{align*}
where $s'_{<i} = \{s'_0, ..., s'_{i-1}\}$ are the tokens preceding $s'_i$ in $s'$. 

We use the same encoder-decoder architecture as T5 \citep{Raffel2019ExploringTL} and initialize our model with pretrained language model weights. The encoder-decoder architecture allows us to perform full attention over the inputs $s,D,q$, and $\mathcal{G}$, while the decoder allows us to auto-regressively generate $s'$. Meanwhile, initializing with pretrained weights has been shown to achieve state-of-the-art results on many NLP tasks \citep{Raffel2019ExploringTL}.

In order to adapt T5 for our task, we represent all our inputs as sequences of tokens. We then concatenate these sequences together using separator tokens, truncating and padding them to fixed lengths. This is straightforward since all our inputs are text. See \cref{fig:model} for reference. We also use the standard cross-entropy loss to train. 
\section{Experiments}\label{sec:experiments}

We train our model on a subset of $\sim$1,020K edits from \dataset. We use a training/validation/test split of 1,000K/10K/10K edits, and train for 3 epochs with a fixed learning rate of 0.0001. Following \citet{Raffel2019ExploringTL}, we finetune all weights in the model, and use a batch size of 128. We validate every 200 steps and select the model with the lowest validation loss.

\subsection{Evaluation}

For inference we use beam search with a beam width of 5, and keep the 5 highest ranked candidates, excluding any generation that parrots the source as this corresponds to making no edits.

\paragraph{Metrics}
We consider several metrics to evaluate our model. One natural metric to consider is BLEU (\cite{Papineni2002BleuAM}). BLEU shows high correlation with human judgement on machine translation \citep{Papineni2002BleuAM, Doddington2002AutomaticEO}. While this should not a priori transfer to evaluating different tasks, our task in fact bears a high similarity to machine translation because of how the output is constrained by the inputs. If, for example, the source sentence in an English to German translation task is ``Sally met Lucy'', the German translation must in some way mention Sally and Lucy. Similarly, in our task, if the source sentence is ``Barack Obama was the 44th President of the United States'', and the command is ``add birth date'', the edit must somehow mention a birth date somewhere. Thus, in our setting, BLEU makes sense as a metric since in principle a good model output should not deviate too far from the reference. We use macro-averaged sentence-level BLEU with epsilon smoothing and equally weighted $n$-grams, with $n$ up to $4$.

One issue with BLEU is that the source and target sentences in our task are already very similar, so a model that simply parrots back the source sentence could achieve an unduly high score. Therefore, we also evaluate model outputs by comparing the word-level edits made by the model against the reference, where a word-level edit is a tuple of an operation, either insertion or deletion, a position, and a word. For example, in the edit ``Barack Obama was the 44\textsuperscript{th} President of the United States'' $\xrightarrow[]{}$ ``Barack Obama, born August 4\textsuperscript{th} 1961, was the 44\textsuperscript{th} President of the United States'', the set of word edits would look like $\{(\text{insert}, 2, ``,"), (\text{insert}, 3, ``\text{born}"), ...\}$. We can then compute precision, recall, and F1 scores based on these word edit sets.\footnote{See \cref{sec:appendix_word_edit_metrics} for a precise description.} 

Finally, we also compute sentence-level accuracy, which reports the number of edits in the test set for which the model output exactly matched the reference.

\paragraph{Baselines}
We use two baselines to compare our model to. First, we consider the parrot baseline that simply outputs the source sentence as is. 
The second baseline attempts to delete the source sentence and replace it with a new sentence. We use a pretrained GPT-2 model \citep{Radford2019LanguageMA} out of the box that generates a sentence given the left context.

\subsection{Results}

\begin{table}[bt]
    \centering
    \scalebox{0.93}{
    \begin{tabular}{@{}l*{5}{r}@{}} \toprule
        Model & Acc. & \multicolumn{3}{c}{Word Edit} & BLEU  \\ \cmidrule(r){3-5}
         & & R & P & F1 & \\ \midrule
\multicolumn{6}{@{}l}{{\it Baselines:}}\\
    Parrot baseline & 0 & 0 & 0 & 0 & \num{0.671865375671978}\\
    GPT-2 & 0 & \num{0.382031689
} & \num{0.046531077
} & \num{0.073362714
} & \num{0.00044784
} \\ 
\midrule
\multicolumn{6}{@{}l}{{\it Ablations:}}\\
Only source & \num{0.169412411
} & \num{0.235995158
} & \num{0.259979203
} & \num{0.230596546
} & \num{0.623791222
}\\ 
No command & \num{0.203368113
} & \num{0.283489266
} & \num{0.311774044
} & \num{0.28048946
} & \num{0.651030362
} \\ 
No grounding & \num{0.180761486362804
} & \num{0.243176760094339
} & \num{0.276406389357623
} & \num{0.242974943910153
} & \num{0.641263227958426
}\\
\midrule
\multicolumn{6}{@{}l}{{\it Our system:}}\\
Interactive & \num{0.302031851
} & \num{0.409097875
} & \num{0.441343591
} & \num{0.406402509
} & \num{0.698329439
} \\ 
\bottomrule
\end{tabular}
}
    \caption{Evaluation of our model (Interactive Editor) against baselines and ablations. The parrot baseline simply outputs the source sentence. The GPT-2 baseline is an out-of-the box GPT-2 model fed only the grounding and left context, and tasked with predicting the next sentence. The ``only source'' system is only fed the source sentence as input.
    }
    \label{tab:main_results}
\end{table}

\Cref{tab:main_results} presents our main results. Notice that the parrot baseline is able to achieve a considerably high BLEU score, as expected, while the GPT-2 baseline surprisingly achieves a high word edit recall score. Our interactive neural editor model is able to beat both baselines across all metrics, as would be expected. Even on a harsh metric like accuracy our model achieves a nontrivial score, although we suspect most of the edits that the model gets exactly right are minor fluency edits. See \cref{tab:intention_breakdown} for a breakdown by edit type.

\paragraph{Ablations} 

\begin{table*}%
    \centering
    \begin{tabular}{p{0.2\linewidth}*{7}{r}} \toprule
        Intention Category & Acc. & \multicolumn{3}{c}{Word Edit} & BLEU  & S-BLEU & \#Edits\\ \cmidrule(r){3-5}
         & & P & R & F1 & \\ \midrule
Fluency & \num{0.363709
} & \num{0.494846
} & \num{0.469727
} & \num{0.464094
} & \num{0.759772
} & \num{0.732819
} & \num{6244
} \\
Content & \num{0.101361
} & \num{0.239635
} & \num{0.194647
} & \num{0.198070
} & \num{0.415474
} & \num{0.380594
} & \num{2792
} \\
Other & \num{0.291378
} & \num{0.449104
} & \num{0.407583
} & \num{0.407824
} & \num{0.738333
} & \num{0.719499
} & \num{3027
} \\
         \bottomrule
    \end{tabular}
    \caption{Breakdown of results by intention category for our full model. The categories are the same as in \cref{tab:data_intention_breakdown_grouped}.}
    \label{tab:intention_breakdown}
\end{table*}

The three middle rows of \Cref{tab:main_results} show the results for three ablations of our model. The first ablation removes everything but the source sentence $s$. This is similar to the paraphrase setting \cite{Gupta2018ADG}, and the editing setting in \citet{Faruqui2018WikiAtomicEditsAM} and \citet{Yin2019LearningTR}. We can see that including the context, grounding, and command as additional inputs yields significant improvements over only using the source sentence. We can also see from the second ablation that the commands are a crucial element in the model's performance. This is not surprising since without a command the model must guess what type of edit to make. While it may make a valid edit, it should very rarely be able to guess the right edit to make. Similarly, the model without grounding performs considerably worse than the full model, showing that the grounding is equally important as the command.
Surprisingly, the last two ablations perform only marginally better than the first, meaning that removing the grounding in addition to the commands, or vice-versa, does not lead to a large drop in performance. This seems to suggest a synergistic effect between the command and the grounding, which makes sense since the model would not know what to do with the grounding without a command, and likewise, the model would not have access to the right information without the grounding, even if it knew what to edit from the command.

\paragraph{Breakdown by edit type}

The results of our full model are broken down by edit intention labels in Table~\ref{tab:intention_breakdown}. The columns report the same metrics as in our main table of results, with the exception of S-BLEU, which reports the BLEU score between the source sentence and target, and the last column, which reports the number of test edits that were classified into each category. With the caveat that intention labels come from an automatic classifier and not human annotation, we can observe that our model has varying performance across different types of edits. The model performs very well on fluency edits, but worse on content edits. This comes at no surprise given that fluency edits should be easier as they usually correct minor mistakes, which a language model should be able to detect from pretraining. Content edits, on the other hand, require pulling the correct information from the grounding and incorporating it in the correct manner into the sentence. The S-BLEU scores confirm this since the source sentences in the fluency examples are much more similar to the target sentences than for the content edits. In fact, when looking at the absolute improvement of the BLEU over the S-BLEU scores, the model performs equally well on both types of edits.

\begin{table}[bt]
    \centering
    \begin{tabular}{@{}l*{3}{r}@{}} \toprule
        Task & \multicolumn{3}{c}{Preference (\%)} \\ 
        \cmidrule{2-4}
         & Reference & Neutral & Interactive\\ \midrule
        Command & 41.00 & 31.71 & 27.29 \\
        Grounding & 29.14 & 34.86 & 36.00 \\
         \bottomrule
    \end{tabular}
    \caption{Human Evaluation: judging preferences for our system (Interactive Editor) versus human references.}
    \label{tab:human_eval}
\end{table}

\subsection{Human Evaluations}
We also conducted human evaluations of our system, comparing our model's top output from beam search to the reference edit across 200 examples from our test set. Annotators were crowd sourced, and each example was rated by 7 judges for a total of 1400 judgments in each of two tasks. 
In the first task, we asked judges to choose which system better accomplished the command $q$. 
In the second, we asked which system was more faithful to the grounding $\mathcal{G}$.  \Cref{tab:human_eval} presents the results. Although there is a  clear preference for the Reference edits in the command-related task, 59\% of judgments suggest that Interactive Editor may be equal to or better than the reference.\footnote{The high percentage of Neutral judgments here may be partially attributable to other factors. Majority Neutral judgments are observed for approximately 65\% of those examples that received at least 1 Neutral judgment. This suggests that many commands may not have been readily interpretable to our judges.} In the grounding task, Interactive Editor demonstrates good correspondence with the background material.
Judges were further asked whether the retrieved grounding was relevant to the context $D$: 92.86\% of judgments recorded the grounding as either "Somewhat relevant" or "Very relevant"

\section{Discussion}

\begin{table}[hbt]
    \centering
    \begin{tabular}{@{}p{0.09\textwidth}p{0.36\textwidth}@{}} \toprule
        Text: & Geoff Hinton is an English tennis player. \\ %
        Command: & {\bf fix profession} \\ %
        Text: & Geoffrey Hinton is a computer science professor at the University of Toronto. \\ %
        Command: & {\bf add nationality} \\ %
        Text: & Geoffrey Hinton is an English-Canadian computer science professor at the University of Toronto. \\ %
        Command: & {\bf add birthdate} \\ %
        Text: & Geoffrey Hinton (born 1946) is an English-Canadian computer science professor at the University of Toronto. \\ %
        Command: & {\bf add most famous work} \\ %
        Text: & Geoffrey Hinton (born 1946) is an English-Canadian computer science professor at the University of Toronto. 
        Geoffrey Hinton is most famous for his work on artificial neural networks.\\
         \bottomrule
    \end{tabular}
    \caption{An example of a multi-turn interaction with our model. At each turn, the edit was chosen among the top 3 outputs returned by beam-search. See \cref{tab:hinton_grounding} in the appendix for the grounding used in this example.}
    \label{tab:hinton_examples}
\end{table}

From our results, our model seems to be able to learn how to make nontrivial edits to text. However we have focused solely on single turns, while a real text generation scenario would likely involve multiple edits. It isn't obvious that a model that performs well on a single turn will also perform well across multiple turns, as there may be path dependencies when making edits. \Cref{tab:hinton_examples} presents an illustrative example of  multi-turn interaction with our model. The starting text is a false sentence about Geoff Hinton, which, through a series of edits, is built into a more elaborate, and factually correct, sentence. The model is able to perform nontrivial reasoning to retrieve relevant information from the grounding and insert it in the appropriate part of the sentence. For example, when asked to ``fix profession'', the model infers that Hinton is not a tennis player, but a computer scientist. See \cref{tab:hinton_grounding} in the appendix for the grounding used in this example. While this example is not an empirical result, it suggests that our single-turn model can be used meaningfully over multiple turns, and could thus potentially be extended for the multi-turn setting.
\section{Related Work}

\paragraph{Grounded Generation} Large language models can generate fluent text \citep{Radford2019LanguageMA, Brown2020LanguageMA, Raffel2019ExploringTL}, but they have a tendency to hallucinate facts \citep{Wiseman2017ChallengesID}. Thus, several works have explored using various forms of grounding to enable models to generate factually consistent texts \citep{KoncelKedziorski2019TextGF, Liu2018TabletotextGB, Prabhumoye2019TowardsCT, Liu2018GeneratingWB, Guu2020REALMRL}. Our work uses grounding to ensure that edits are factually correct, although our task differs from previous work because of the user command, which requires specific information to be retrieved from the grounding during generation.

\paragraph{Controllable Generation} While grounding can be seen as a way to implicitly control the contents of generated text, other works have explored more explicit forms of control. \citet{Hokamp2017LexicallyCD} and \citet{Zhang2020POINTERCT} use lexical constraints, while \citet{Keskar2019CTRLAC} and \citet{Dathathri2020PlugAP} control higher level attributes of text, such as style, tone, or topic. Our task instead uses natural language commands, which can flexibly express different types of constraints, ranging from low-level lexical ones, to high-level topical ones. In this sense, we can also draw the parallel to dialog response generation \citep{Ghazvininejad2018AKN, Dinan2019WizardOW}, task-oriented dialog \citep{Gao2019NeuralAT}, or open domain question answering \citep{Min2019KnowledgeGT,Chen2017ReadingWT}, that also involve 
user responses or queries, although these tasks are not concerned with text generation in the context of document creation.

\paragraph{Story Generation}
The task of Document Generation considered in our work bears similarity with work on generating long-form narratives
\cite{jain2017story}. While earlier work in Story Generation 
focused more on plan-based architectures \cite{Lebowitz1985StorytellingAP}, more recent work moved towards end-to-end approaches 
\cite{Fan2018HierarchicalNS} allowing generation to be unconstrained and creative. As narratives are often aimed at particular goals expressed in terms of outlines and plans, much of the literature in Story Generation is framed as a form of controllable generation, using 
storylines \cite{Peng2018TowardsCS},
events \cite{martin2017event,Harrison2017TowardAS},
plot words or word skeletons \cite{xu-etal-2018-skeleton,Ippolito2019UnsupervisedHS},
plans \cite{Yao2019PlanAndWriteTB}, 
story ending \cite{Tambwekar2019ControllableNS},
and outlines \cite{rashkin2020plotmachines} 
as various forms of constraints.
Our work takes a significantly different approach, as we 
treat document or story generation as an iterative process that also allows a human to generate a full document from scratch, but allows constraints to be more ``on demand'' and dynamic (e.g., express the desire to add nationality in Table~\ref{tab:hinton_examples} only if the system missed that the first time).

\paragraph{Text Editing} Several previous works have focused on text editing. \citet{Guu2018GeneratingSB} generate sentences by editing prototypes taken from their training corpus, although they use editing only as a means for language modeling. \citet{Wu2019ResponseGB} expand upon \citet{Guu2018GeneratingSB}'s setting, but for dialog. More related to our own setting, \citet{Faruqui2018WikiAtomicEditsAM} propose WikiAtomicEdits, a dataset of edits crawled from Wikipedia. However, they consider a much narrower definition of edits than our data does. \citet{Yin2019LearningTR} uses WikiAtomicEdits and proposes the task of learning to represent edits, which \citet{MarreseTaylor2020VariationalIF} expands using a variational approach. In contrast, we are more interested in generating edits rather than representing them. \citet{Iso2020FactbasedTE} propose a fact-based text editing task, but they do not consider control or other types of edits. Another related task to text editing is text paraphrasing \citep{Gupta2018ADG}, however paraphrasing usually conserves the meaning of a sentence. While the edits we consider include meaning-preserving edits, such as rewording edits, we are mostly interested in edits that add or modify content.
\section{Conclusion}
In this work we argued that text generation should be interactive, and, as a means towards that end, we proposed a general text editing task, where a system must edit a document in response to a user command. In our specific instance of the task we considered single-sentence edits, and we crawled a dataset of several million edits from Wikipedia that included commands, in the form of editor comments, as well as grounding documents. We then showed that training a transformer-based model on our data, while initializing with pretrained language model weights, yields encouraging results on both automatic and human evaluations. Additionally, our ablation studies showed the crucial role played by the user command and grounding. Breaking down our results by types of edits, we saw that our model not only performs well on easier fluency edits, but also on much harder content edits. Finally, our example of multi-turn interactions with the system suggests that our single-turn editing model could be feasibly used to generate longer documents.

\bibliography{references}
\bibliographystyle{acl_natbib}

\appendix

\begin{table*}
    \centering
    \begin{tabular}{@{}p{0.2\linewidth} p{0.5\linewidth}r r@{}} \toprule
         Label & Description & \%Edits & \%Orig. \\ \midrule
         Counter-Vandalism & Revert or otherwise; remove vandalism & \num{0.05467169646274124} & \num{1.9} \\
         Fact-update & Update numbers, dates, scores, episodes, status, etc. based on newly available information & \num{1.5722017854214017} & \num{5.5} \\
         Copy-editing & Rephrase; improve grammar, spelling, tone, or punctuation & \num{29.218116638159284} & \num{11.8}\\
         Wikification & Format text to meet style guidelines, e.g. add links or remove them where necessary & \num{21.124362488968032} & \num{33.1} \\
         Vandalism & Deliberately attempt to damage the article & \num{1.0075212633848027} & \num{2.5} \\
         Simplification & Reduce the complexity or breadth of discussion; may remove information & \num{3.1272210376687988} & \num{1.6} \\
         Elaboration & Extend/add substantive new content; insert a fact or new meaningful assertion & \num{9.500378796754063} & \num{12} \\
         Verifiability & Add/modify references/citations; remove unverified text & \num{7.634511898904223} & \num{5.4} \\
         Process & Start/continue a wiki process workflow such as tagging an article with cleanup, merge or deletion notices & \num{0.6240383639104321} & \num{4.4} \\
         Clarification & Specify or explain an existing fact or meaning by example or discussion without adding new information & \num{3.5388208096097222} & \num{0.7} \\
         Disambiguation & Relink from a disambiguation page to a specific page & \num{0.7005787389582698} & \num{0.3} \\
         Point-of-view & Rewrite using encyclopedic, neutral tone; remove bias; apply due weight & \num{0} & \num{0.3} \\ 
         Unlabeled & No label & \num{21.389129704694735}  & \num{1.2} \\
         \bottomrule
    \end{tabular}
    \caption{Breakdown of the edits in our data by intention label. The descriptions are taken from \citet{Yang2017IdentifyingSE}. \%Edits gives the prevalence of each label in our data, while \%Orig. gives the prevalence in the hand-labelled dataset presented in \citet{Yang2017IdentifyingSE}. The percentages do not total $100$ because edits can have multiple labels.}
    \label{tab:data_intention_breakdown}
\end{table*}

\begin{table*}%
\begin{adjustwidth}{-.0in}{-.0in}
    \centering
    \scalebox{0.96}{
    \begin{tabular}{@{}p{.2\linewidth}p{.8\linewidth}@{}} \toprule
        Comment & Reword \\ \cmidrule(r){2-2}
        Source & ByteDance responded by adding a kids-only mode to TikTok which \textbf{allows music} videos \textbf{to be recorded, but not posted} and \text{by removing some accounts} and content \textbf{from those determined to be underage}. \\ \cmidrule(r){2-2}
        Target & ByteDance responded by adding a kids-only mode to TikTok which \textbf{blocks the upload of} videos, \textbf{the building of user profiles, direct messaging,} and \textbf{commenting on other's videos, while still allowing the viewing} and \textbf{recording of} content. \\
        \midrule
        Comment & corrected tense for decedent \\ \cmidrule(r){2-2}
        Source & While Bob Steward \textbf{has} not \textbf{been} an active producer since 1992, he \textbf{serves} as a Creative Consultant in his son's new production company, Steward Television, and \textbf{is} listed on the official website as Steward Television's founder. \\ \cmidrule(r){2-2}
        Target & While Bob Steward \textbf{was} not an active producer since 1992, he \textbf{served} as a Creative Consultant in his son's new production company, Steward Television, and \textbf{was} listed on the official website as Steward Television's founder. \\
        \midrule 
        Comment & fixed spelling for Walter Yetnikoff \\ \cmidrule(r){2-2}
        Source & Mottola was hired by Sony Music ( then known as CBS Records ) by its controversial President Walter \textbf{Yentlkoff} to run its U.S. operations. \\ \cmidrule(r){2-2}
        Target & Mottola was hired by Sony Music ( then known as CBS Records ) by its controversial President Walter \textbf{Yetnikoff} to run its U.S. operations. \\ 
        \bottomrule
    \end{tabular}
    }
    \caption{More example edits from \dataset. The edited portions are highlighted in bold.}
    \label{tab:appendix_edit_examples}
\end{adjustwidth}
\end{table*}

\begin{table*}[hbt]
\begin{adjustwidth}{-.0in}{-.0in}
    \centering
    \begin{tabular}{p{\textwidth}}
    \toprule
    Geoffrey Everest Hinton CC FRS FRSC (born 6 December 1947) is an English Canadian cognitive psychologist and computer scientist, most noted for his work on artificial neural networks.Since 2013 he divides his time working for Google (Google Brain) and the University of Toronto.In 2017, he cofounded and became the Chief Scientific Advisor of the Vector Institute in Toronto. Geoffrey Hinton : index. Department of Computer Science : email: {\tt [REDACTED]} : University of Toronto : voice: send email: 6 King's College Rd. We would like to show you a description here but the site won’t allow us. Geoffrey’s great grandfather, the mathematician {\tt [REDACTED]} Charles Hinton, coined the word “tesseract” and popularized the idea of higher dimensions, while his father, Howard Everest Hinton, was a distinguished entomologist. Geoffrey Hinton is a fellow of the Royal Society, the Royal Society of Canada, and the Association for the Advancement of Artificial Intelligence. He is an honorary foreign member of the American Academy of Arts and Sciences and the National Academy of Engineering, and a former president of the Cognitive Science Society. Geoffrey Hinton. Emeritus Prof. Comp Sci, U.Toronto \& Engineering Fellow, Google. Verified email at cs.toronto.edu - Homepage. machine learning psychology artificial intelligence cognitive science computer science. Articles Cited by Co-authors. Title. Sort. Sort by citations Sort by year Sort by title. Geoff Hinton was born in Wimbledon in 1947 to Howard Hinton, an entomologist, and a schoolteacher mother, Margaret Clark. The childhood Hinton describes is a mash-up of Lemony Snicket, ... As the first of this interview series, I am delighted to present to you an interview with Geoffrey Hinton. Welcome Geoff, and thank you for doing this interview with deeplearning.ai. $\rangle\rangle$ Thank you for inviting me. $\rangle\rangle$ I think that at this point you more than anyone else on this planet has invented so many of the ideas behind deep learning. Talks by Geoffrey Hinton. The next generation of neural networks A 45min version of this talk which was given at the 10 year celebration of the Microsoft Cambridge Research Laboratory. the original powerpoint file version for most browsers.ps version with 4 slides per page. Very gentle after-dinner version of IJCAI-2005 Research Excellence ...\\
    \bottomrule
    \end{tabular}
    \caption{Grounding used for the example in \cref{tab:hinton_examples}. Parts indicated by {\tt [REDACTED]} were removed for containing sensitive material.}
    \label{tab:hinton_grounding}
\end{adjustwidth}
\end{table*}

\section{Appendices}

\subsection{Data Processing pipeline}
\label{sec:appendix_data_pipeline}
This section describes our pipeline to obtain atomic edits from Wikipedia revisions in more detail.
We start by filtering the revisions in the data. In particular, following \cite{zhang2019ModelingTR}, we only keep revisions that affect a single section, and we exclude revisions that do not contain an editor comment. We also exclude certain page types like talk or user pages.

We then strip the Wikipedia markup in the retrieved text, using the WikiExtractor script \citep{Wikiextractor2015}. This removes most markup and Wikimedia templates from the text. Because the markup language used on Wikipedia is not completely formalized\footnote{See \url{https://www.mediawiki.org/wiki/Markup_spec} for a discussion.}, and because malformed markup often appears in intermediate versions of Wikipedia pages, there is no guarantee that we can remove all the markup from the text.

We then split each section into sentences using the Punkt sentence tokenizer \citep{Kiss2006UnsupervisedMS} provided in the NLTK python package \citep{Bird2009NaturalLP}.

After splitting into sentences, we attempt to match the sentences from the pre-edit (source) document to the sentences in the post-edit (target) document. Unmatched sentences will be candidates for edits. Similarly to \cite{Faruqui2018WikiAtomicEditsAM}, for each sentence $s_i$ in the source document, we only look at the target sentences $\{t_{i-k},...,t_i,...,t_{i+k}\}$, with $k=20$. This avoids the quadratic complexity of looking at all matches.

We then filter out revisions that contain more than one sentence-level edit to ensure that the comment is relevant. If there is a single unmatched source, respectively target, sentence, we consider it a sentence deletion, respectively insertion. Because we do not look at all matches between source and target sentences, a sentence may remain unmatched if, in the target document, it was moved more than $k$ sentences away compared to the source document. Thus we only keep a sentence insertion or deletion if the total number of source and target sentences differ by one. If there are both an unmatched source sentence $s$ and target sentence $t$, we consider them to form an edit $s\xrightarrow{}t$ if $f(s,t)>\epsilon$, where $f$ is the BLEU score and $\epsilon=0.1$.

As a final step, we filter out edits that involve sentences with markup punctuation. We have found that this helps remedy the shortfalls of the markup removal step, since it often leaves behind markup symbols. While there may be valid sentences that use markup punctuation, we do not expect them to make up a significant part of the data, nor do we expect them to be significantly different from regular sentences, except for their use of unusual punctuation.

\subsection{Grounding Search Query Construction}
\label{sec:appendix_grounding_search_keywords}
For a given edit, we combine the relevant page and section titles with keywords from the target sentence to construct a query that we use to retrieve grounding from a commercial search engine. In order to identify keywords we look at document frequency
$$\text{df}(w)=\frac{|\{D\in\mathcal{D}\,|\,w\in D\}|}{|\mathcal{D}|},$$
where $\mathcal{D}$ is a sample of $500,000$ Wikipedia articles taken from the Tensorflow Wikipedia dataset (\citenum{wikidump}). We consider words $w$ with $\text{df}(w)<0.01$ to be keywords.

\subsection{Grounding Document Reranking}
\label{sec:appendix_grounding_reranking}
Because the combined length of the grounding snippets we retrieve far exceeds the capacity of our model, we rerank the retrieved snippets using an information extraction score. We then merge the ranked snippets and take only the first $N=512$ tokens. 
Following \cite{Liu2018GeneratingWB} we use tf-idf scores to rerank. For a given edit $s\xrightarrow{}s'$, with retrieved grounding documents $\mathcal{G}$, the information extraction score of snippet $G\in\mathcal{G}$ is
$$\text{score}(G) = \sum_{w\in s'}\text{tf-idf}(w, G),$$
where the tf-idf score of word $w$ is
$$\text{tf-idf}(w, G)=N_w(G)\cdot\text{log}\left(\frac{N_g}{N_{gw}}\right),$$ where $N_w(G)$ is the number of occurrences of $w$ in $G$, $N_{gw}$ is the number of documents in $\mathcal{G}$ that contain $w$, and $N_{g}$ is the number of documents in~$\mathcal{G}$.

\subsection{Word Edit Metrics}
\label{sec:appendix_word_edit_metrics}
For our evaluations we compare the word-level edits made by the model against the reference, where a word-level edit is a tuple of an operation, either insertion or deletion, a position, and a word. We use the word tokenizer from NLTK \citep{Bird2009NaturalLP} to break a sentence down into words. For a given target sentence $s'$, denote the set of word edits as $\text{WE}(s',s)$. Then we compute the precision
$$P_{\text{WE}}(s',h,s)=\frac{|\text{WE}(s',s)\cap \text{WE}(h,s)|}{|\text{WE}(h,s)|},$$
recall,
$$R_{\text{WE}}(s',h,s)=\frac{|\text{WE}(s',s)\cap \text{WE}(h,s)|}{|\text{WE}(s',s)|},$$
and F1 score,
$$F_{1,\text{WE}}(s',h,s) = 2\cdot\frac{P_{\text{WE}}\cdot R_{\text{WE}}}{P_{\text{WE}} + R_{\text{WE}}},$$
where $s$ is the source sentence, $s'$ is the reference target sentence and $h$ is the target sentence generated by the model.

\end{document}